\documentclass[journal]{IEEEtran}
\usepackage[font=small,skip=1pt]{caption}
\IEEEoverridecommandlockouts
\usepackage{textcomp}
\usepackage{gensymb}
\usepackage{latexsym}
\usepackage[T1]{fontenc}
\usepackage{multirow}
\usepackage{graphicx}
\usepackage{varioref}
\usepackage{array}
\usepackage{enumitem}
\usepackage{soul,color}
\usepackage{longtable}
\usepackage{algorithmic}
\usepackage[ruled, vlined]{algorithm2e}
\usepackage{pifont}
\usepackage{booktabs}

\ifCLASSOPTIONcompsoc
  \usepackage[caption=false,font=normalsize,labelfont=sf,textfont=sf]{subfig}
\else
  \usepackage[caption=false,font=footnotesize]{subfig}
\fi

\usepackage{color}
\definecolor{light}{rgb}{0.5, 0.5, 0.5}

\usepackage{graphicx}
\medskip
\usepackage{amssymb}
\usepackage{amsmath,epsfig,amssymb}
\usepackage{amsthm}
\usepackage{xcolor}
\usepackage[active]{srcltx} 
\usepackage{epstopdf} 

\usepackage{bm}
\usepackage{amsthm}
\usepackage{enumitem}

\usepackage{cite}
\usepackage{amsmath,amssymb,amsfonts}
\usepackage{algorithmic}
\usepackage{graphicx}
\usepackage{textcomp}
\usepackage{xcolor}
\def\BibTeX{{\rm B\kern-.05em{\sc i\kern-.025em b}\kern-.08em
    T\kern-.1667em\lower.7ex\hbox{E}\kern-.125emX}}
\usepackage{lineno,hyperref}
\usepackage[utf8]{inputenc}
\usepackage[english]{babel}
\usepackage{mathdots}

\usepackage[utf8]{inputenc}
\usepackage[T1]{fontenc}

\usepackage{graphicx}
\usepackage{float}
\usepackage{array, tabularx, booktabs, threeparttable}
\usepackage{multirow}
\usepackage{subfig}
\usepackage{longtable}
\usepackage{subcaption} 

\usepackage{xcolor}
\usepackage{soul}

\usepackage[ruled, vlined]{algorithm2e} 

\usepackage{amsmath, amssymb, bm, mathdots}
\usepackage{amsthm}
\usepackage{varioref}
\usepackage{gensymb}
\usepackage{pifont}
\usepackage{textcomp}

\usepackage[font=small, skip=1pt]{caption}

\usepackage{cite}

\usepackage{hyperref}
\usepackage[utf8]{inputenc}

\graphicspath{{Pictures/}{../jpeg/}} 
\usepackage{float}
\usepackage{amsmath}

\begin{document}

\title{Multi-Lingual Cyber Threat Detection in Tweets/X Using ML, DL, and LLM: A Comparative Analysis}

\author{\IEEEauthorblockN{Saydul Akbar Murad\textsuperscript{1}, Ashim Dahal\textsuperscript{1}, and Nick Rahimi\textsuperscript{1}}\\
\IEEEauthorblockA{\textsuperscript{1}\textit{School of Computing Sciences and Computer Engineering}\\
\textit{University of Southern Mississippi}
Hattiesburg, MS, 39406, USA\\
\{saydulakbar.murad, ashim.dahal, nick.rahimi\}@usm.edu}
}






\maketitle
\begin{abstract}
Cyber threat detection has become an important area of focus in the digital age of today due to the growing spread of fake information and harmful content on social media platforms such as Twitter (now 'X'). These cyber threats, often disguised within tweets, pose significant risks to individuals, communities, and even nations, emphasizing the need for effective detection systems. While previous research has explored tweet-based threats, much of the work is limited to specific languages, domains, or locations or relies on single-model approaches, reducing their applicability to diverse real-world scenarios. To address these gaps, our study focuses on multi-lingual tweet cyber threat detection using a variety of advanced models. The research was conducted in three stages: (1) We collected and labeled tweet datasets in four languages—English, Chinese, Russian, and Arabic—employing both manual and polarity-based labeling methods to ensure high-quality annotations. (2) Each dataset was analyzed individually using ML and DL models to assess their performance on distinct languages. (3) Finally, we combined all four datasets into a single multi-lingual dataset and applied DL and LLM architectures to evaluate their efficacy in identifying cyber threats across various languages. Our results show that among machine learning models, Random Forest (RF) attained the highest performance; however the Bi-LSTM architecture consistently surpassed other DL and LLMs architecture across all datasets. These findings underline the effectiveness of Bi-LSTM in multilingual cyber threat detection. The code of this paper can be found in this link: \href{https://github.com/Mmurrad/Tweet-Data-Classification.git}{https://github.com/Mmurrad/Tweet-Data-Classification.git}
\end{abstract} 

\section{Introduction}

Social networks have emerged as a global focal point, functioning as platforms for the dissemination of opinions, ideas, marketing, and several other activities \cite{r2}. These platforms offer customers rapid, complimentary, and readily available tools to fulfill their varied requirements. Social networks efficiently fulfill user needs, resulting in a continuing rise in account registrations. Users continuously express their opinions on subjects of interest, rendering social networks a vibrant arena for communication and engagement. Twitter, now renamed as 'X,' is recognized as one of the most prominent and impactful social networks. Tweets have emerged as a significant data source for many users, including manufacturers, celebrities, healthcare experts, politicians, and researchers \cite{r4}. The extensive information dissemination on Twitter makes it a significant asset for activities such as cyber threat identification, sentiment analysis, and predictive modeling, underscoring the influence of social networks in contemporary society.

With the increase of social media, cyber threats have become an increasing concern, affecting both individuals and organizations in different ways \cite{c8}. threats come in various forms, such as online harassment, phishing scams, misinformation, and direct threats, using anonymity and etc. \cite{c9}. Some threats are sent privately through messages that target specific people, while others are made public. The goal is spreading fear among people or manipulating public opinion \cite{c1}. These types of cyberbullying can affect human life, leading to emotional distress, misinformation can create chaos or ruin reputations, and phishing scams can cause financial losses \cite{c5}. As social media increasingly embedded in our daily lives, so we need an effective systems to identify and mitigate these threats, ensuring a safer and more secure digital platform for everyone. 

In recent years, cyber threats based on tweets have become a growing concern, especially through the exploitation of public tweets \cite{c1}. Such threats can result in serious consequences, often leading to public unrest or even community violence \cite{c2}. Fake or misleading tweets can sometimes incite mobs, creating panic and chaos in society \cite{c5}.  On a broader level, such tweets can exacerbate group differences, inciting hostility along cultural, religious, or ideological lines \cite{c6}. In many instances, aggressive or misleading tweets have instigated diplomatic conflicts between nations, thereby straining international relations \cite{c7}. The rapid dissemination of harmful content on platforms such as Twitter underlines the urgent necessity to address these challenges and protect societal and global stability.

Many researchers are actively working on detecting cyber threats in tweets, using advanced AI techniques such as ML \cite{r7}, DL \cite{r8}, and other novel approaches. Although significant progress has been made, there are still several limitations in the current body of research. One major issue is that most of the research concentrates on a single language \cite{r10}. But everyday tweets are coming in different languages. This language-specific focus limits the scalability and applicability of the models in addressing threats in different linguistic contexts. Some researchers worked on multi-lingual tweet datasets, but mostly their approaches involve devloping separate models for each language \cite{bengali}. This approach may be effective for small-scale investigations, but it becomes wasteful and impracticable when used for big datasets comprising multiple languages. To address the diversity of cyber threats on social media, it's important to develop a generalized model capable of classifying tweets across multiple languages effectively.

Another challenge is performance. Many researchers are struggling to deliver robust results, particularly when handling complex datasets with diverse linguistic \cite{c4}. Moreover, the absence of thorough comparisons among various architectures—such as ML, DL, and LLM-based models—makes it difficult to identify the most effective approach for tackling cyber threats in tweets \cite{c3}. Without such comparisons, it is ambiguous which approaches provide the optimal equilibrium of accuracy, scalability, and flexibility. Considering these limitations, it is important to develop a unified, generalized model to handle multi-lingual tweet datasets. Additionally, a thorough evaluation of different model architectures is essential to understand their strengths and weaknesses to build more efficient and reliable systems. Addressing these gaps will not only advance the field of cyber threat identification but also foster a safer, more secure online world.


To address the limitations of previous research, we focused on identifying cyber-threats in tweet data across multiple languages. Our study employed in two different ways. In the first approach, we applied ML and DL models separately to each language dataset—English, Chinese, Russian, and Arabic—to evaluate their performance on individual datasets. In the second approach, we combined all four language datasets into a unified, multi-lingual dataset and used DL and LLMs for classification. The threat identification task involved multi-class categorization for English, Chinese, and Russian tweets, while the Arabic dataset was treated as a binary classification problem. In the single-language analysis, both ML and DL models demonstrated strong performance, effectively identifying threats within individual datasets. However, when working with the combined dataset, the performance of the models was less satisfactory, with results falling short of expectations. As this project is ongoing, we are actively working to improve the performance of the models on the combined dataset. The contributions of our research are as follows:


\begin{itemize}
    \item We independently collected four diverse datasets of tweets containing potential cyber threats, covering English, Chinese, Russian, and Arabic, to ensure a comprehensive and multi-lingual analysis.  

    \item The datasets were labeled using a combination of manual annotation and polarity-based methods, specifically designed to identify cyber threat-related content, with cross-validation to ensure high accuracy.  

    \item Experiments were conducted in two approaches: analyzing each dataset individually to evaluate model performance on single-language tweet threat detection and combining all datasets for a multi-lingual tweet threat analysis.  

    \item  We employed three modeling techniques—ML, DL, and LLMs—to thoroughly compare their effectiveness in detecting tweet threats across diverse datasets.  

\end{itemize}

These contributions address gaps in previous research and provide a robust framework for tweet classification across languages.

The remainder of this research article is organized as follows: Section \ref{literature} provides an overview of related work, highlighting key studies and comparing them with our approach. Section \ref{meth} outlines the methodology, detailing the data preprocessing steps, the models used, and the performance metrics employed in the analysis. Section \ref{res} presents and discusses the experimental results. Finally, Section \ref{con} concludes the paper by summarizing the findings and discussing potential directions for future work.

\section{Literature Review}\label{literature}

This section reviews the current state-of-the-art (SOTA) research on multi-lingual threat detection on Twitter/X. Although ML, DL, and LLMs have widely reached the depths of multiple disciplines and achieved commendable results, multi-lingual threat detection ceases to remain one of them. This gap is particularly concerning given the current rising prevalence of cross-country cyber threats and social media based cyber attacks.

Rehan et al. \cite{english-urdu} claimed to be one of the first to offer multi-lingual threatening text detection on Twitter with LLMs. They achieved this by first translating English text into their Urdu corpus and then working on the Urdu language with their AI implementation. The authors fine-tuned RoBERTa with 1,313 English and 2,400 Urdu samples. The sample of English threats is highly skewed with only 128 non-threat messages out of the given 1,313 samples. The authors also chose standard ML algorithms like Support Vector Machine (SVM), Logistic Regression (LR), Random Forest (RF), Convolutional Neural Networks (CNN), Bi-directional Long Short-Term Memory (Bi-LSTM) with RoBERTa and Word2Vec approaches to test their approach. Although they have shown exceptional results with over 91.89\% accuracy, the definition of multi-lingual classification for the papers includes just English and Urdu. Our paper overcomes this shortcoming by providing a more quantifiable and diverse approach that can work in English, Arabic, Russian, and Chinese; all of which, according to the CIA's The World Factbook \cite{cia_factbook}, are in the top 10 most-spoken first languages around the world. This limitation in detecting multiple language to detect cyber threat can lead to catastrophic cyber attack vectors where the exploiters simply make use of the standing language and its essence barriers.

Apart from this, there exists very little literature on multi-language threat detection on Twitter tweets specifically. However, there are some related research works in the field which indirectly address the theme of this paper. Tundis et al. \cite{berlin_bomb} provide a multi-language approach towards identification of suspicious users on social network platforms. Although their approach also relies on using platforms like Google Translate, Yandex, and Bing Translate, the authors deduce a similarity score. The authors don't mention a general formula to calculate the similarity score, but we can deduce it to the following from their examples.

\begin{equation}
\begin{aligned}
    S_{N_{max}} = maxSimilarity(&\sum_{i=1}^{n} (R_{1i} \times S_{1i}); \\
    &\sum_{i=1}^{n} (R_{2i} \times S_{2i}); \\
    &\sum_{i=1}^{n} (R_{3i} \times S_{3i}))
\end{aligned}
\label{eq:max_similarity}
\end{equation}

where $S_{ij} \in [0,1]$, $i$ and $j$ $\in \{1,2,3\}$, is the similarity score of all tweets between the 3 service providers, (i.e, $S_{12}$ denotes similarity between service 1 and service 2) and $ R_{ij} \in {0,1}$ is the binary indication that gives if the similarity score of $S_{ij}$ is associated with the service or not.

From evaluation of Equation \ref{eq:max_similarity}, it can be evident that the function grows exponentially as i and j grow, i.e., adding more service translators for languages that may not be best performed by the given three translators. We again argue for the need of a language-dependent corpus system because some essence of the original message could be lost within translation. Furthermore, the authors only apply naive approaches like Bag of Words (BoW) alongside bi-gram and tri-gram versions of N-gram technique. This provides a more statistical analysis of whether the profile of some users could be dangerous or not rather than a prediction with a finely labeled dataset and advanced ML and DL algorithms. We overcome this drawback by using a finely labeled dataset which gives robust prediction against tweets that contain cyber threats.

\begin{table*}[ht]
\renewcommand{\arraystretch}{1.3}
\caption{Summary of Multi-lingual Threat Detection/Analysis Studies}
\scriptsize
\label{tab:literature-summary}
\centering
\begin{tabular}{l ccc c ccc cccc c}
\hline
\multirow{2}{*}{\textbf{Study}} & 
\multicolumn{3}{c}{\textbf{Data Source}} & 
\multirow{2}{*}{\textbf{\begin{tabular}[c]{@{}c@{}}Number of\\Classes\end{tabular}}} & 
\multicolumn{3}{c}{\textbf{Models Used}} & 
\multicolumn{4}{c}{\textbf{Metrics}} & 
\multirow{2}{*}{\textbf{\begin{tabular}[c]{@{}c@{}}Content Language\end{tabular}}} \\
\cline{2-4} \cline{6-12}
& \textbf{Tweet} & \textbf{FB} & \textbf{IG} & 
& \textbf{ML} & \textbf{DL} & \textbf{LLM} & 
\textbf{Acc.} & \textbf{Prec.} & \textbf{Rec.} & \textbf{F1} & \\
\hline
\cite{english-urdu} & \checkmark & \ding{55} & \ding{55} & 2 & SVM, LR, RF & CNN, BiLSTM & RoBERTa & \checkmark & \checkmark & \checkmark & \checkmark & English, Urdu \\
\cite{berlin_bomb} & \checkmark & \checkmark & \checkmark & 2 & BoW, NGram & \ding{55} & \ding{55} & \checkmark & \ding{55} & \ding{55} & \ding{55} & Multi-lingual \\
\cite{french_hate_speech} & \checkmark & \checkmark & \ding{55} & 2 & SVD & BERT & \ding{55} & \checkmark & \checkmark & \checkmark & \ding{55} & English, French \\
\cite{multiclass_classification} & \checkmark & \ding{55} & \ding{55} & 4 & \begin{tabular}[c]{@{}c@{}}SVM, RF\\LR, DT, NB\end{tabular} & \ding{55} & \ding{55} & \checkmark & \checkmark & \ding{55} & \ding{55} & English \\
\cite{llm_content_analysis} & \checkmark & \checkmark & \ding{55} & Multi & \ding{55} & \ding{55} & LLaMA & \checkmark & \ding{55} & \ding{55} & \ding{55} & English, Hindi, Arabic \\
\cite{sentiment_multilingual} & \checkmark & \ding{55} & \ding{55} & 3 & \ding{55} &  BERT & RoBERTa, GPT-3 & \checkmark & \ding{55} & \ding{55} & \ding{55} & Multi-lingual \\
\cite{offensive_behaviour} & \checkmark & \checkmark & \ding{55} & 5 & \ding{55}& \ding{55} & XLM-RoBERTa  & \checkmark & \ding{55} & \ding{55} & \ding{55} & Bengali \\
Our work & \checkmark & \ding{55} & \ding{55} & Multi & LR, DT, RF & RNN, LSTM, GRU & XLM-RoBERTa  & \checkmark & \checkmark & \checkmark & \ding{55} & Eng., Russian, Chinese, Arabic \\
\hline
\end{tabular}
\begin{tablenotes}

\item[a] \textbf{Abbreviations:} X = Twitter, FB = Facebook, IG = Instagram, ML = Machine Learning (SVM = Support Vector Machine, LR = Logistic Regression, RF = Random Forest, DT = Decision Tree, NB = Naive Bayes, BoW = Bag of Words, NGram = N-gram, SVD = Singular Value Decomposition), DL = Deep Learning (CNN = Convolutional Neural Network, BiLSTM = Bidirectional Long Short-Term Memory, BERT = Bidirectional Encoder Representations from Transformers, XLM-RoBERTa = Cross-lingual RoBERTa), LLM = Large Language Model (GPT-3 = Generative Pre-trained Transformer 3).
\end{tablenotes}
\end{table*}

In other similar works, Chiril et al. \cite{french_hate_speech} comparatively studied and experimented with several methods and models to detect multi-lingual hate speech towards multiple targets. Their dataset contains 13,071 English and 3,085 French tweets classified into hate and non-hate tweets towards immigrants and women. They achieve precision of 0.78 and 0.66 in the two tasks respectively. The authors, however, do not train on more variety or advanced techniques like LLMs or GRUs. The benefit of not implementing a translation layer for multi-language system can be observed in this study as even with simple techniques like dimensionality reduction and singular value decomposition the authors were able to get good results in their dataset. Hussein et al. \cite{multiclass_classification} did multi-class classification of tweets on Threat (8,280), Business (2,331), Irrelevant (6,598) or Unknown (4,159); number of samples in brackets. They experimented with SVM, Random Forrest (RF), Logistic Regression (LR), Decision Tree (DT), Naive Bayes (NB), and K-Nearest-Neighbor (KNN) algorithms and got the best result with RF classification with a precision of 74 and accuracy of 67. Similar to the previous literature, we conclude we could use more advanced methods and models to bridge the research gap presented by the findings of this article as well.

In the context of analyzing text for cybersecurity threats, cyberbullying or hate speech by using SoTA LLMs, Kmainasi et al. \cite{llm_content_analysis} fine tuned Llama to LlamaLens which is a multilingual LLM for analyzing news and social media content. Their dataset comtains ~2.7 million samples and over 222 labels, all of which is an amalgamation of 103 dataset consisting multiple social media post, news article, political debates and transcripts. Out of the three language tested: English, Hindi and Arabic, cyberbullying was the only common category of interest in all three languages where the authors gained accuracy of 90.07, 60.90 and 86.30 respectively. Hindi and Arabic had two common hate speech categories. Offensive speech was yet another common category in all three of them, but the definition of the term itself could be ambiguous.  Miah et al. \cite{sentiment_multilingual} worked with ensemble learning of transformer and LLM for multi-lingual sentiment analysis. Although they work on five languages: Arabic, Chinese, English, French and Italian, their method depend upon a translation layer as that of \cite{english-urdu,berlin_bomb}. The author's ensemble consisted of Twitter‑Roberta‑Base‑Sentiment‑Latest, bert‑base‑multilingual‑uncased‑sentiment, and GPT‑3 and produced an accuracy of 86\% on all languages. Although the authors argue that sentiment analysis is possible through translation to English given their ensemble method achieved the accuracy, but it's not the effective way to do so. We show that by using a language specific corpus and model we can handle nuanced information within the language like slang, irony, and sarcasm away from the actual threats present in the dataset.

Hirdi et al. \cite{offensive_behaviour} utilized different BERT based method to detect offensive behaviour on low resource language Bengali. They trained XLM-RoBERTa-base with 44,000 comments from social media platforms and gained an accuracy of 83.54\%. This accuracy is lower than \cite{sentiment_multilingual}, but the authors worked with a different language with not as many resource for accurate translation and also accounted for Banglish which is a combination of Bengali and English; either Bengali written in romanized English form or a mix of both language into a single sentence. This practice of writing is common in many foreign language system which don't use the roman alphabet as English alphabets can be used to produce similar sounds as the other characters in Arabic, Chinese, Devanagari or Bengali scripts \cite{devenglish}\cite{devenglish1}\cite{bengali}. The authors divided the remarks in the 44,000 comments into one of five categories: Sexual, Not bully, Troll, Religious, Threat. The authors employed technique to translate English words in a Bengali text to Bengali to fit for this corpus which is a better approach than translating the entire text from one language into another. The author's work in the Bengali language is commendable. Their work highlights the importance of language understanding by constructing methodology for Bengali which is not written in the roman alphabet and use the finer details observed to determine the level of threat present in the given tweet/post. We do argue that their approach could be expanded to cover bases for multiple languages which our research touches upon.

Table \ref{tab:literature-summary} shows a details comparison between our work and previous research. Our work adds contribution to existing literature and offers a deeper research into classifying tweets for safer online interaction by incorporating a comprehensive analysis of all ML, DL and LLM algorithms in 4 of the top 5 most commonly spoken first language.

\section{Implementation Details}\label{meth}

\begin{figure*}[t]
    \centering
    \includegraphics[width=1\linewidth]{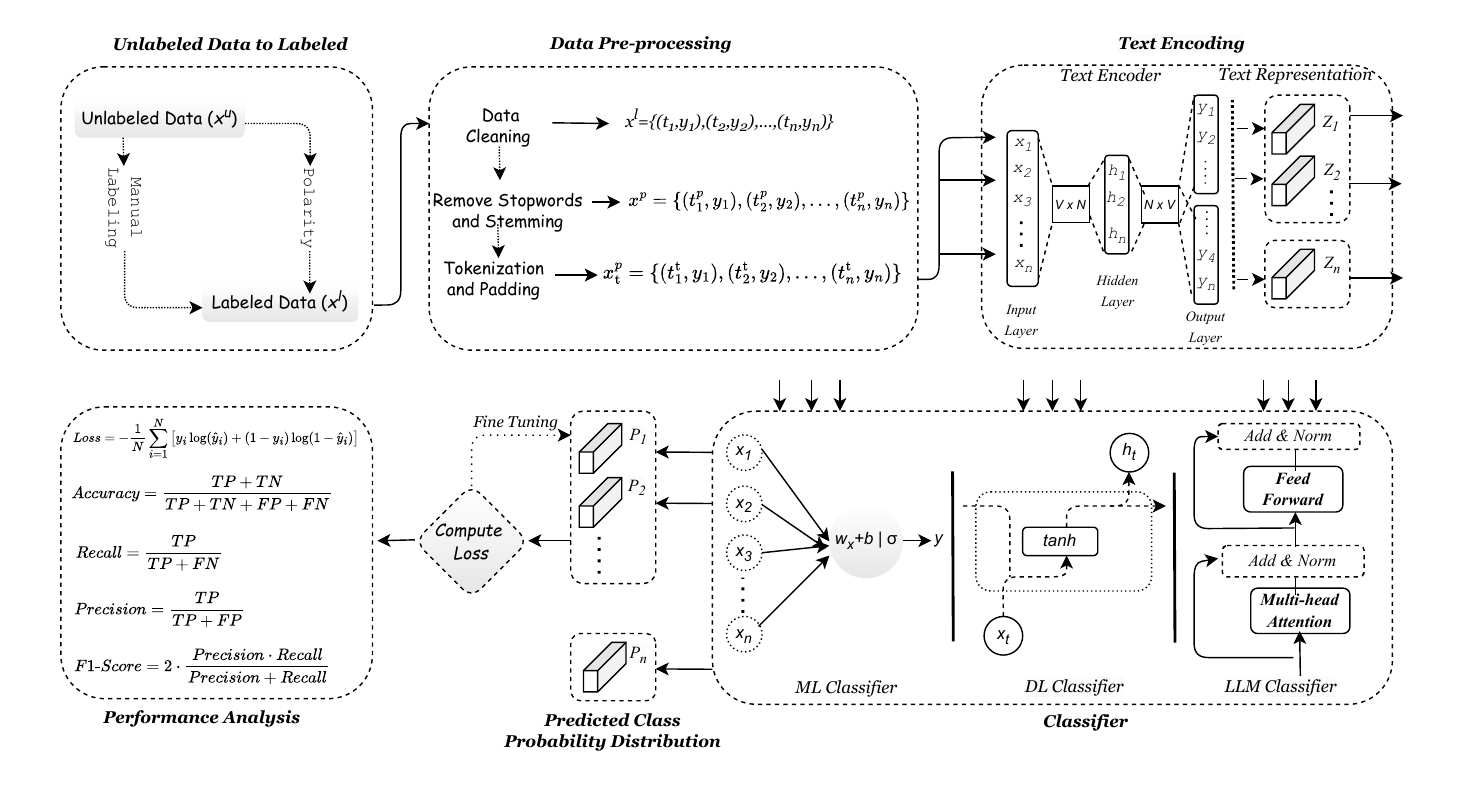}
    \caption{Workflow of the Tweet Data Classification Process.}
    \vspace{-10pt}
    \label{arch}
\end{figure*}

In this section, we briefly discuss the working procedure of this research. We started by discussing the data, including its source and the process of labeling it. Next, we explained how we processed the data, which included steps such as data cleaning, removing stopwords, and tokenization. After processing, we encoded the tokenized data and applied various techniques to the dataset. Finally, we concluded the section by describing the different parameters considered to measure the performance of the models. Figure \ref{arch} provides a visual representation of our working procedure.

\subsection{Data Collection and Labeling}
For this research, we collected our dataset from tweets posted by individuals across different languages. The dataset comprises tweets in four languages: English, Arabic, Russian, and Chinese. Initially, the collected data was unlabeled, and the objective was to classify the tweets into three categories: threat, neutral, and non-threat. We have made the dataset publicly available, and it's included with the source of this paper. 

We employed two labeling approaches to annotate the dataset. The first approach was manual annotation, where four students, each proficient in one of the four languages, assisted in classifying the tweets into the predefined categories. This ensured linguistic nuances were considered during the labeling process.

The second approach is polarity-based classification. Polarity $(P)$ measures the sentiment of a tweet and was calculated as follows:

Given a tweet $T$ consisting of $n$ words $w_1, w_2, \ldots, w_n$, the Polarity of the tweet is calculated as:

\begin{equation}
    P(T)=\frac{\sum_{i=1}^n s\left(w_i\right)}{n}
\end{equation}

Where $P(T)$ is the polarity of the tweet. $s\left(w_i\right)$ represents the sentiment score of the $i$-th word in the tweet. This score is positive for positive words, negative for negative words, and zero for neutral words. $n$ is the total number of words in the tweet.

To classify tweets based on their polarity:

\begin{itemize}
    \item Threat: $P(T) \leq-0.5$
    \item Neutral: $-0.5<P(T)<0.5$
    \item Non-threat: $P(T) \geq 0.5$ 
\end{itemize}

The polarity-based labeling approach yielded results that were largely consistent with the manual labeling. However, in certain instances, discrepancies were observed between the two methods. In such cases, we prioritized the manual labeling as the final annotation, given its reliability and context-sensitive accuracy. This combined approach ensured the dataset was comprehensively and accurately labeled.

\subsection{Data Pre-processing}
In the data preprocessing stage, we began by cleaning unnecessary data to enhance the quality of the dataset and improve the performance of the classification models. Following this, we removed stopwords from the tweets in all four languages (English, Arabic, Russian, and Chinese). Additionally, stemming was performed to reduce words to their root forms, ensuring consistency and further simplifying the data. Finally, the cleaned text data was tokenized, breaking the tweets into individual words or subwords, and padding was applied to standardize the input length. These steps prepared the data for model training, ensuring compatibility and optimal input representation for the subsequent stages of analysis.

\subsubsection{Data Cleaning}
After labeling the datasets, our first objective was to clean the data to ensure it was suitable for analysis. Many tweets contained sentences that lacked meaningful context, so we removed such sentences. Additionally, we eliminated URLs, mentions, hashtags, punctuation, and special characters using Regular Expressions (Regex).

We denote the labeled dataset as:

\begin{equation}
X^l=\left\{\left(t_1, y_1\right),\left(t_2, y_2\right), \ldots,\left(t_n, y_n\right)\right\}
\end{equation}

where $t_i$ represents the $i$-th tweet. $y_i$ is the corresponding label for $t_i$, where $y_i \in\{$ Threat, Neutral, Non-threat $\}$.

The cleaning process transforms each tweet $t_i$ into a cleaned version $t_i^{\prime}$ by applying the following operations:
1. Removal of irrelevant or nonsensical sentences.
2. Removal of URLs, mentions, hashtags, punctuation, and special characters via Regex.

The cleaned dataset is represented as:

\begin{equation}
X_{\text {c }}^l=\left\{\left(t_1^{\prime}, y_1\right),\left(t_2^{\prime}, y_2\right), \ldots,\left(t_n^{\prime}, y_n\right)\right\}
\end{equation}

Where $t_i^{\prime}$ is the cleaned version of $t_i$. This ensures that the dataset $X_{\text {clean }}^l$ retains the labels $y_i$ while improving the quality and relevance of the tweets $t_i^{\prime}$.

\subsubsection{Remove Stopwords and Stemming}
After completing the data cleaning process, we proceeded to remove stopwords from the tweets in all four languages. Stopwords, which are common words that do not significantly contribute to the semantic meaning of the text. These were eliminated to reduce noise and enhance the quality of the dataset. This step can be expressed mathematically as:

$$
t_i^{\prime \prime}=\text { RemoveStopwords }\left(t_i^{\prime}\right)
$$

Where $t_i^{\prime}$ is the cleaned tweet from the previous stage. $t_i^{\prime \prime}$ is the tweet after removing stopwords. RemoveStopwords represents the stopword removal function applied to the cleaned tweet.

After removing stopwords, stemming was applied to reduce words to their root or base forms, further standardizing the text. This transformation is represented as:

$$
t_i^{\prime \prime \prime}=\operatorname{Stem.}\left(t_i^{\prime \prime}\right)
$$

Where $t_i^{\prime \prime \prime}$ is the tweet after stemming. Stem $(\cdot)$ represents the stemming function.

The resulting dataset after these transformations is represented as:

\begin{equation}
X_{\text {s }}^l=\left\{\left(t_1^{\prime \prime \prime}, y_1\right),\left(t_2^{\prime \prime \prime}, y_2\right), \ldots,\left(t_n^{\prime \prime \prime}, y_n\right)\right\}
\end{equation}

This ensures that the processed tweets $t_i^{\prime \prime \prime}$ retain their associated labels $y_i$, making the dataset ready for tokenization and further preparation for model training. 

\subsubsection{Tokenization and Padding}
The final steps of our data preprocessing pipeline were tokenization and padding, essential processes for preparing the text data for machine learning models.

\paragraph{\textbf{Tokenization}}
Tokenization involves breaking down the text into smaller units and converting these units into numerical values. For this research, we set a vocabulary limit of \( \text{max\_words} = 5000 \), meaning we only retained the 5000 most frequent words in the dataset. Words outside this vocabulary were replaced with a special token indicating "out of vocabulary." 

Mathematically, we can represent a preprocessed tweet \( t_i''' \) as a tokenized sequence:

\begin{equation}
T_i = \text{Tokenize}(t_i''')
\end{equation}

Here \( t_i''' \) is the cleaned tweet after stopword removal and stemming. \( T_i = \{w_1, w_2, \dots, w_k\} \) is the sequence of tokenized words. Each token \( w_j \) is an integer index ranging from 1 to \( \text{max\_words} \), corresponding to its place in the vocabulary.

\paragraph{\textbf{Padding}}
Since tweets can vary in length, padding ensures consistency by standardizing the length of all tokenized sequences to \( \text{maxlen} = 500 \). If a sequence is shorter than 500 tokens, zeros are added (padding). If it is longer, it is truncated to the first 500 tokens.

The padded version of a tokenized tweet \( T_i \) is:

\begin{equation}
    P_i = \text{Pad}(T_i, \text{maxlen})
\end{equation}

Where \( T_i \) is the tokenized tweet, \( P_i = \{p_1, p_2, \dots, p_{\text{maxlen}}\} \) is the padded sequence, \( p_j = w_j \) for \( j \leq k \), and \( p_j = 0 \) for \( j > k \) when the sequence is shorter than 500 tokens.

After tokenization and padding, the final dataset is represented as:

\begin{equation}
    X^l_{\text{t}} = \{(P_1, y_1), (P_2, y_2), \dots, (P_n, y_n)\}
\end{equation}

Where \( P_i \) is the padded sequence of the \( i \)-th tweet and \( y_i \) is the corresponding label.

This standardized approach significantly improved the data quality and ensured compatibility with the subsequent analytical processes.

\subsection{Text Encoding}

After preprocessing, the next step was encoding the text data into numerical vector representations using Word2Vec. Word2Vec transforms words into dense, continuous vector spaces where semantically similar words have closer representations. 

For training Word2Vec, we utilized different pre-trained word embedding models specific to each language:

\begin{itemize}
    \item English: GoogleNews-vectors-negative300.bin (300-dimensional vectors)
    \item Chinese: glove.840B.300d.txt (300-dimensional vectors)
    \item Russian: tweets\_model.w2v (custom model)
    \item Arabic: A Word2Vec model trained on our dataset.
\end{itemize}

\subsection{Word2Vec Representation}
Each word \( w \) in a tweet \( t_i''' \) was mapped to a vector \( \vec{w} \) in a \( d \)-dimensional space, where \( d \) is the embedding size (300 in this case). Mathematically, for a tweet \( t_i''' = \{w_1, w_2, \dots, w_k\} \), the Word2Vec embeddings create a sequence of word vectors:

\begin{equation}
    E_i = \{\vec{w}_1, \vec{w}_2, \dots, \vec{w}_k\}, \quad \vec{w}_j \in \mathbb{R}^d
\end{equation}

\paragraph{Aggregating Word Embeddings}
As ML models typically require fixed-length inputs, the sequence of word embeddings for a tweet \( E_i \) was aggregated into a single vector. Common aggregation techniques include:

\textbf{Mean Pooling:} Taking the average of all word vectors:

$$\vec{T}_i = \frac{1}{k} \sum_{j=1}^k \vec{w}_j$$

\textbf{Max Pooling:} Taking the maximum value across each dimension of the word vectors:

    $$\vec{T}_i = \max(\vec{w}_1, \vec{w}_2, \dots, \vec{w}_k)$$

Here, \( \vec{T}_i \in \mathbb{R}^d \) represents the encoded tweet as a fixed-length vector.

\paragraph{Passing Encoded Data to the Model}
Once encoded, the tweet representations \( \vec{T}_i \) were used as input to the classification model. The final dataset after encoding is represented as:

\begin{equation}
    X^l_{\text{en}} = \{(\vec{T}_1, y_1), (\vec{T}_2, y_2), \dots, (\vec{T}_n, y_n)\}
\end{equation}

Where \( \vec{T}_i \) is the vector representation of the \( i \)-th tweet, \( y_i \) is the corresponding label.

These embeddings preserve semantic information and ensure that the text data is represented in a numerical format suitable for training machine learning or deep learning models. The use of language-specific pre-trained models further enhanced the quality of the encoded representations by leveraging linguistic and contextual knowledge inherent in each model. 

\subsection{Classifier}

To analyze the dataset, we employed three distinct types of models: ML, DL, and LLMs. For the ML approach, we utilized three robust algorithms: Logistic Regression (LR), Decision Tree (DT), and Random Forest (RF), known for their effectiveness in classification tasks. In the DL category, we designed architectures combining advanced recurrent neural networks, including LSTM, and GRU, to take advantage of their sequential data processing capabilities. These models provided a comprehensive framework for evaluating and classifying the dataset.

\subsubsection{\textbf{Language-Specific Classification}}
In the experiment, we used ML and DL models for each language—English, Chinese, Russian, and Arabic. The primary goal was to evaluate the performance of those models when trained and tested exclusively on data from a single language. To achieve this, we employed three ML models and three DL architectures, and we compared the results between the models.

\paragraph{\textbf{ML Classifier}}

For training the ML models, we utilized 80\% of the dataset, while the remaining 20\% was reserved for testing. Each algorithm was carefully configured with hyperparameters optimized for the classification task. In the following, we detail the three ML algorithms employed:

\textbf{Logistic Regression (LR):} The LR model maps the input features \( \vec{T}_i \) to probabilities using a sigmoid function. The predicted probability of a label \( y_i \) is given by:

\begin{equation}
    P(y_i = 1 | \vec{T}_i) = \frac{1}{1 + e^{-(\vec{w}^\top \vec{T}_i + b)}}
\end{equation}

Where, \( \vec{w} \) is the weight vector, \( b \) is the bias term.

We trained the LR model with a maximum of 1000 iterations (\( \text{max\_iterations} = 1000 \)) to ensure convergence.

\textbf{Decision Tree (DT):} The Decision Tree algorithm partitions the feature space \( \vec{T}_i \) into regions by recursively splitting the data based on features that maximize information gain or minimize Gini impurity. The decision rule at each node can be expressed as:

\begin{equation}
    f(\vec{T}_i) = 
\begin{cases} 
1 & \text{if } T_{ij} \leq \text{threshold} \\
0 & \text{otherwise}
\end{cases}
\end{equation}

Where \( T_{ij} \) is the \( j \)-th feature of \( \vec{T}_i \) and \( \text{threshold} \) is the splitting value determined during training.

\textbf{Random Forest (RF):}
Random Forest is an ensemble algorithm that builds multiple Decision Trees (\( DT_1, DT_2, \dots, DT_M \)) on random subsets of the data and features. The final prediction is obtained by majority voting:

\begin{equation}
    f(\vec{T}_i) = \text{mode}\left(DT_1(\vec{T}_i), DT_2(\vec{T}_i), \dots, DT_M(\vec{T}_i)\right)
\end{equation}

Where \( M \) is the number of trees in the forest, and \( DT_m(\vec{T}_i) \) is the prediction of the \( m \)-th tree.

Each algorithm was trained on the encoded dataset \( X^l_{\text{encoded}} = \{(\vec{T}_1, y_1), (\vec{T}_2, y_2), \dots, (\vec{T}_n, y_n)\} \), where \( \vec{T}_i \) represents the encoded tweet and \( y_i \) the corresponding label. By using distinct approaches such as linear modeling (LR), hierarchical splits (DT), and ensemble learning (RF), we ensured diverse perspectives in analyzing and classifying the dataset.

\paragraph{\textbf{DL Classifier}}

\begin{figure*}[t]
    \centering
    \includegraphics[width=1\linewidth]{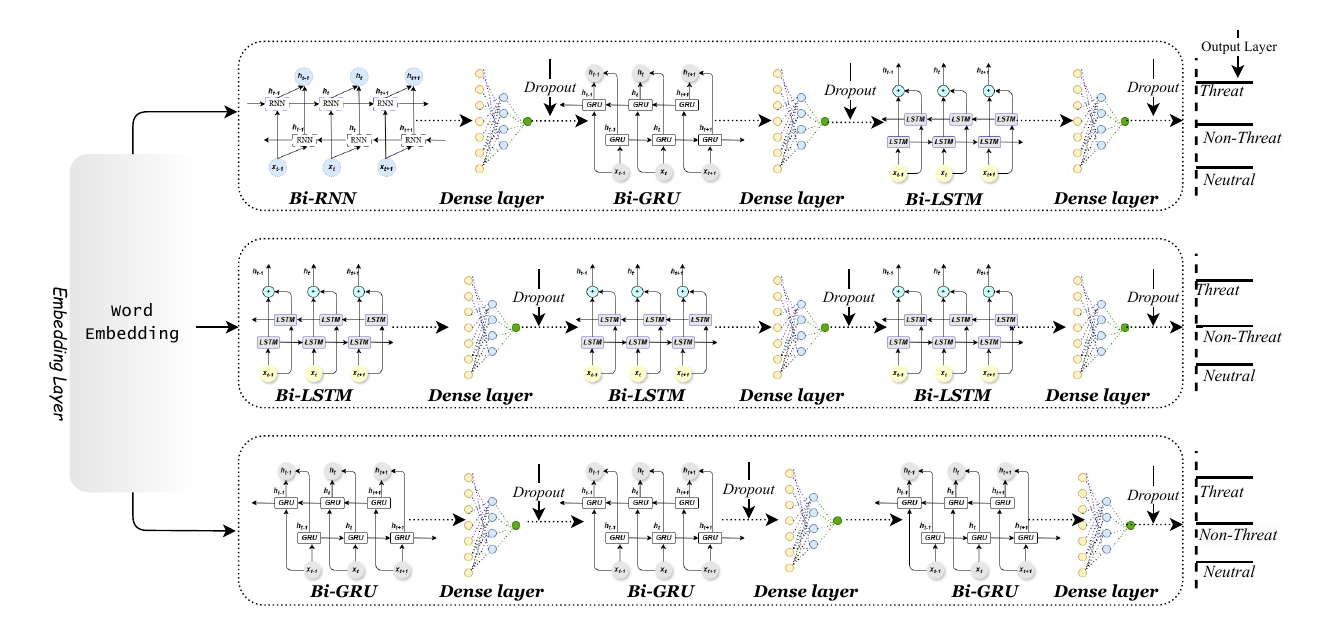}
    \caption{Workflow of the Tweet Data Classification Process.}
    \vspace{-10pt}
    \label{method}
\end{figure*}

We employed three distinct DL architectures Bi-RNN, Bi-LSTM, and Bi-GRU to analyze the dataset. Each architecture was designed and tested independently to evaluate its performance and identify the most effective model. Additionally, we meticulously tuned the hyperparameters of each architecture to achieve optimal results. A brief overview of these models and their configurations, as used in our experiments, is presented in Figure \ref{method}. Below, we provide detailed descriptions of the models and configurations used in our experiments.

\textbf{BiDirectional Recurrent Neural Framework (Bi-RNF)}
The top architecture in Figure \ref{method} defines the RNF models, which begin with a word embedding layer common to all other architectures. The word embedding layer transforms the input tokens into dense vectors of dimension \( d = 300 \), with a maximum sequence length of 100. The embedding layer is initialized with pre-trained embeddings. 

The first layer after embedding is a Bi-RNN with different units like 32, 64, 128, configured to return sequences. This layer captures temporal dependencies in both forward and backward directions. Mathematically, the output of this layer is:

   $$H_{\text{BiRNN}} = \text{BiRNN}(E) $$

where \( E \) is the embedding matrix for a sequence, and \( H_{\text{BiRNN}} \) represents the bidirectional output.

Following the Bi-RNN layer, the output is passed through a dense layer with different neurons and a ReLU activation function:

    $$D_1 = \text{ReLU}(W_{\text{Dense1}} H_{\text{BiRNN}} + b_{\text{Dense1}})$$

where \( W_{\text{Dense1}} \) and \( b_{\text{Dense1}} \) are the weights and biases of the dense layer.

To prevent overfitting, a dropout layer with a rate of 0.4 is applied:

   $$D_1^{\text{drop}} = \text{Dropout}(D_1, 0.4)$$

This process is repeated for subsequent layers, including a Bidirectional GRU (Bi-GRU) and a Bidirectional LSTM (Bi-LSTM), interspersed with dense layers (with varying neuron counts such as 32, 64, or 128) and dropout layers. Each bidirectional layer captures sequential patterns, and the dense layers act as feature transformation layers, enhancing the learning capacity. 

The output from the final dense layer is passed to another dense layer with \( \text{num\_classes} \) neurons and a softmax activation function, producing a probability distribution for multi-class classification:

\begin{equation}
    P(y | T) = \text{Softmax}(W_{\text{Output}} H_{\text{Dense}} + b_{\text{Output}})
\end{equation}

Here, \( W_{\text{Output}} \) and \( b_{\text{Output}} \) are the weights and biases of the final dense layer, \( H_{\text{Dense}} \) is the output from the last dense layer, and \( P(y | T) \) represents the predicted probabilities.

The model was compiled using the Adam optimizer with a learning rate of 0.005 and sparse categorical cross-entropy as the loss function. The architecture ensures a robust framework for processing sequential data, using bidirectional layers to capture comprehensive context and dense layers to enhance representational power.

\textbf{BiDirectional Long Short-term Memory (Bi-LSTM)}
The second architecture in Figure \ref{method} is the Bi-LSTM model, which also begins with the same word embedding layer as the Bi-RNF architecture, transforming input sequences into dense vectors of dimension \( d = 300 \). The first layer after embedding is a Bi-LSTM with 32 neurons, configured to return sequences. For a sequence \( E \) from the embedding layer, the output of this layer is:

    $$H_{\text{BiLSTM1}} = \text{BiLSTM}(E)$$

Following the Bi-LSTM layer, the output is passed through a dense layer with different neurons and a ReLU activation function:

   $$ D_1 = \text{ReLU}(W_{\text{Dense1}} H_{\text{BiLSTM1}} + b_{\text{Dense1}}) $$

where \( W_{\text{Dense1}} \) and \( b_{\text{Dense1}} \) are the weights and biases of the dense layer. To prevent overfitting, a dropout layer with a rate of 0.4 is applied:

   $$D_1^{\text{drop}} = \text{Dropout}(D_1, 0.4)$$

This process is repeated for two other Bi-LSTM layers, each configured to return sequences, followed by dense layers with different neuron counts (e.g., 128) and dropout layers. The output for the second Bi-LSTM layer is:

    $$H_{\text{BiLSTM2}} = \text{BiLSTM}(D_1^{\text{drop}})$$

The subsequent dense and dropout layers are applied as follows:

  $$ D_2 = \text{ReLU}(W_{\text{Dense2}} H_{\text{BiLSTM2}} + b_{\text{Dense2}}) $$

$$ D_2^{\text{drop}} = \text{Dropout}(D_2, 0.4) $$

The final Bi-LSTM layer is set to return a single sequence instead of a complete sequence:

   $$ H_{\text{BiLSTM3}} = \text{BiLSTM}(D_2^{\text{drop}}) $$

Before generating the output, the architecture implements a final dropout layer with a rate of 0.4:

  $$  H_{\text{BiLSTM3}}^{\text{drop}} = \text{Dropout}(H_{\text{BiLSTM3}}, 0.4) $$

The output layer comprises a dense layer with \( \text{num\_classes} \) neurons and employs a softmax activation function for multi-class classification.

\begin{equation}
    P(y | T) = \text{Softmax}(W_{\text{Output}} H_{\text{BiLSTM3}}^{\text{drop}} + b_{\text{Output}})
\end{equation}

where \( W_{\text{Output}} \) and \( b_{\text{Output}} \) denote the weights and biases of the output layer, while \( P(y | T) \) denotes the projected probability distribution for the classes. This architecture uses the sequential learning capabilities of three Bi-LSTM layers, interspersed with dense and dropout layers, to extract contextual information and deliver accurate multi-class predictions.

\textbf{BiDirectional Gated Recurrent Unit (Bi-GRU):}
The final architecture in Figure \ref{method} is the Bi-GRU model, which closely follows the structure of the Bi-LSTM architecture described earlier. The primary difference lies in replacing the Bi-LSTM layers with Bi-GRU layers. GRU is a simplified variant of LSTM that uses fewer parameters while maintaining similar performance for sequence modeling tasks. 

The Bi-GRU model starts with the same word embedding layer, where input tokens are transformed into dense vectors of dimension \( d = 300 \). The first layer after embedding is a Bi-GRU with 128 neurons, configured to return sequences. For a sequence \( E \) from the embedding layer, the output of the initial Bi-GRU layer is:

  $$  H_{\text{BiGRU1}} = \text{BiGRU}(E) $$

This is followed by a dense layer with 32 neurons and a ReLU activation function:

  $$  D_1 = \text{ReLU}(W_{\text{Dense1}} H_{\text{BiGRU1}} + b_{\text{Dense1}}) $$

To alleviate overfitting, a dropout layer is applied with rate 0.4:

  $$  D_1^{\text{drop}} = \text{Dropout}(D_1, 0.4) $$

The architecture contains more Bidirectional GRU layers, each designed to process the output from the previous layer, using differing neuron counts (e.g., 64 and 32). Regarding the second Bi-GRU layer:

   $$ H_{\text{BiGRU2}} = \text{BiGRU}(D_1^{\text{drop}}) $$

Upon passing the third Bi-GRU layer, designed to output a singular sequence instead of a whole sequence:

  $$  H_{\text{BiGRU3}} = \text{BiGRU}(H_{\text{BiGRU2}}) $$

Finally, the model employs a dense layer including \( \text{num\_classes} \) neurons, utilizing a softmax activation function to provide output probabilities for multi-class classification.

\begin{equation}
    P(y | T) = \text{Softmax}(W_{\text{Output}} H_{\text{BiGRU3}} + b_{\text{Output}})
\end{equation}

This design effectively captures sequential dependencies through the gating mechanism of GRUs, while demonstrating greater computational efficiency than Bi-LSTM. 

\subsubsection{\textbf{Multi-class Classifier}}
Before implementing the multi-class classification models, we combined datasets from all four languages—English, Chinese, Russian, and Arabic—into a single unified dataset. Each dataset was processed independently up to the embedding stage, ensuring that the word embeddings were created separately for each language while maintaining a consistent embedding dimension across all datasets. After generating embeddings, we merged the four embedded datasets into one.

It is worth noting that the datasets for English, Chinese, and Russian contained three class labels (Threat, Neutral, Non-threat), while the Arabic dataset included only two labels (Threat and Non-threat). Despite this difference, we preserved the same embedding dimension across all datasets to ensure uniformity and compatibility during the model training process. This unified, multi-lingual dataset provided the foundation for training and evaluating the multi-class classification models, including the LLM-based classifiers. 

\paragraph{\textbf{DL Classifier}}
For the multi-class classification task, we utilized two distinct DL architectures: LSTM and GRU. The architecture of those models is same that we already discussed earlier. To maximize efficiency and improve outcomes, we modified the hyperparameters to better align with the integrated multilingual dataset. This methodology enabled us to evaluate the efficacy of these models while modifying them to address the complexities of multi-class classification in various languages.

\paragraph{\textbf{LLM Classifier}}
For the multi-class classification task, we also used an LLM architecture, specifically XLM-RoBERTa. The objective was to evaluate its performance in comparison to the DL architectures, such as LSTM and GRU. The technique utilized the strong multilingual capabilities of XLM-RoBERTa, which was pre-trained on a vast corpus of text in many languages, rendering it very appropriate for the integrated multilingual dataset.

The LLM architecture begins with tokenization via the XLM-RoBERTa tokenizer, transforming input text into numerical representations while maintaining uniformity in sequence length by padding and truncation. For a dataset \( D = \{t_1, t_2, \dots, t_n\} \), the tokenized input is represented as:

\[
T = \text{Tokenize}(D, \text{max\_len})
\]

where \( \text{max\_len} = 100 \) is the maximum sequence length.

The tokenized data \( T \) is then passed to the XLM-RoBERTa model, which outputs the last hidden state \( H \):

\[
H = \text{XLM-RoBERTa}(T)
\]

Here, \( H \) is a tensor of size \( (n, \text{max\_len}, d) \), where \( d \) is the hidden state dimension of the model. To reduce this tensor to a fixed-size vector for each input sequence, we apply a global average pooling operation:

\[
P = \text{GlobalAveragePooling}(H)
\]

The pooled output \( P \) serves as the input to a series of dense layers. The first dense layer transforms \( P \) with 128 neurons and ReLU activation:

\[
D_1 = \text{ReLU}(W_1 P + b_1)
\]

Dropout is applied after each dense layer to mitigate overfitting:

\[
D_1^{\text{drop}} = \text{Dropout}(D_1, 0.4)
\]

This process continues for two more dense layers with 64 and 32 neurons, respectively, each followed by dropout layers. The final dense layer, configured with \( \text{num\_classes} \) neurons and a softmax activation function, produces the output probabilities:

\begin{equation}
    P(y | T) = \text{Softmax}(W_{\text{Output}} D_3^{\text{drop}} + b_{\text{Output}})
\end{equation}

where \( P(y | T) \) represents the probability distribution over the class labels for the input text.

The model was trained with a learning rate of \( 0.0005 \), a batch size of 16, and for 50 epochs. It was evaluated using sparse categorical cross-entropy loss and accuracy metrics. 

\subsection{Performance Analysis}

To evaluate the performance of all models, we considered different parameters, including loss, accuracy, precision, recall, and F1-score. These metrics provide a comprehensive view of models performance for all four tweet datasets—English, Chinese, Russian, and Arabic. Each metric takes a specific aspect of the models' performance, allowing for a detailed comparison of their strengths and limitations. The mathematical definitions of these performance metrics are presented in Figure \ref{method}. These equations form the foundation for our analysis, illustrating how the metrics were calculated to ensure consistent and accurate evaluation of the models.

\section{Result}\label{res}

In this experiment, we utilized ML, DL, and LLM techniques to analyze tweet threats in four different languages: English, Chinese, Russian, and Arabic. We completed the experiment in two different ways. In the first approach, we applied ML and DL models to each language dataset separately; that's allowed us to evaluate performance on individual datasets. In the second approach, we combined all four datasets into a single multi-lingual dataset and utilized DL and LLM models to assess their performance.

\begin{table*}[]
\scriptsize
\caption{Performance Comparison of ML and DL Models for English Cyber Tweet Threat Detection}
\label{tab2_english}
\begin{tabular}{@{}ccccccccccccccccccc@{}}
\toprule
\multicolumn{2}{c}{\multirow{2}{*}{Model}} & \multirow{2}{*}{Params} & \multicolumn{3}{c}{\begin{tabular}[c]{@{}c@{}}Precision Per Category \\ in \%\end{tabular}} & \multicolumn{3}{c}{\begin{tabular}[c]{@{}c@{}}Recall Per Category \\ in \%\end{tabular}} & \multicolumn{3}{c}{\begin{tabular}[c]{@{}c@{}}F1-Score Per Category \\ in \%\end{tabular}} & \multirow{2}{*}{Accuracy} & \multirow{2}{*}{PMA} & \multirow{2}{*}{RMA} & \multirow{2}{*}{FMA} & \multirow{2}{*}{PWA} & \multirow{2}{*}{RWA} & \multirow{2}{*}{FWA} \\ \cmidrule(lr){4-12}
\multicolumn{2}{c}{}                       &                         & Th                           & Neu                          & Non-Th                        & Th                          & Neu                         & Non-Th                       & Th                           & Neu                         & Non-Th                        &                           &                      &                      &                      &                      &                      &                      \\ \cmidrule(r){1-19}  
\multirow{3}{*}{ML}        & LR            & \multirow{3}{*}{}       & 0.19                         & 0.63                         & 0.66                          & 0.45                        & 0.30                        & 0.68                         & 0.26                         & 0.41                        & 0.67                          & 0.53                      & 0.49                 & 0.48                 & 0.45                 & 0.59                 & 0.53                 & 0.54                 \\
                           & DT            &                         & 0.54                         & 0.70                         & 0.80                          & 0.50                        & 0.70                        & 0.80                         & 0.52                         & 0.70                        & 0.80                          & 0.74                      & 0.68                 & 0.67                 & 0.67                 & 0.74                 & 0.74                 & 0.74                 \\
                           & RF            &                         & \textbf{0.96}                         & 0.75                         & 0.78                          & 0.43                        & 0.67                        & \textbf{0.91}                         & \textbf{0.59}                         & 0.71                        & 0.84                          & \textbf{0.78}                      & \textbf{0.83}                 & 0.67                 & 0.71                 & 0.79                 & \textbf{0.78}                 & 0.77                 \\ \cmidrule(l){1-19}
\multirow{3}{*}{DL}        & Bi-RNF        & 22.67K                  & 0.21                         & 0.00                         & 0.94                          & \textbf{0.98}                       & 0.00                        & 0.80                         & 0.34                         & 0.00                        & 0.87                          & 0.56                      & 0.38                 & 0.60                 & 0.40                 & 0.54                 & 0.56                 & 0.52                 \\
                           & Bi-LSTM       & 23.64K                  & 0.36                         & \textbf{0.83}                         & 0.95                          & 0.73                        & 0.75                        & 0.82                         & 0.49                         & \textbf{0.79}                        & \textbf{0.88}                          & \textbf{0.78}                      & 0.71                 & \textbf{0.77}                 & \textbf{0.72}                 & \textbf{0.85}                 & \textbf{0.78}                 & \textbf{0.80}                 \\
                           & Bi-GRU        & 26.05K                  & 0.39                         & 0.64                         & \textbf{0.96}                          & 0.46                        & \textbf{0.83}                        & 0.75                         & 0.42                         & 0.72                        & 0.84                          & 0.75                      & 0.66                 & 0.68                 & 0.66                 & 0.79                 & 0.75                 & 0.76                 \\ \cmidrule(l){1-19} 
\end{tabular}
\begin{tablenotes}
\item[a] \textbf{Abbreviations:} ML = Machine Learning, DL = Deep Learning, LR = Logistic Regression, DT = Decision Tree, RF = Random Forest, Bi-RNF = Bidirectional Recurrent Neural Framework, Bi-LSTM = Bidirectional Long Short-Term Memory, Bi-GRU = Bidirectional Gated Recurrent Unit, Th = Threat, Neu = Neutral, Non-Th = Non-Threat, PMA = Precision Macro Average, RMA = Recall Macro Average, FMA = F1-Score Macro Average, PWA = Precision Weighted Average, RWA = Recall Weighted Average, and FWA = F1-Score Weighted Average.
\end{tablenotes}
\end{table*}

\begin{table*}[]
\scriptsize
\caption{Performance Comparison of ML and DL Models for Chinese Cyber Tweet Threat Detection}
\label{chinese_result}
\begin{tabular}{@{}ccccccccccccccccccc@{}}
\toprule
\multicolumn{2}{c}{\multirow{2}{*}{Model}} & \multirow{2}{*}{Params} & \multicolumn{3}{c}{\begin{tabular}[c]{@{}c@{}}Precision Per Category \\ in \%\end{tabular}} & \multicolumn{3}{c}{\begin{tabular}[c]{@{}c@{}}Recall Per Category \\ in \%\end{tabular}} & \multicolumn{3}{c}{\begin{tabular}[c]{@{}c@{}}F1-Score Per Category \\ in \%\end{tabular}} & \multirow{2}{*}{Accuracy} & \multirow{2}{*}{PMA} & \multirow{2}{*}{RMA} & \multirow{2}{*}{FMA} & \multirow{2}{*}{PWA} & \multirow{2}{*}{RWA} & \multirow{2}{*}{FWA} \\ \cmidrule(lr){4-12}
\multicolumn{2}{c}{}                       &                         & Th                            & Neu                          & Non-Th                       & Th                           & Neu                         & Non-Th                      & Th                           & Neu                          & Non-Th                       &                           &                      &                      &                      &                      &                      &                      \\ \cmidrule(r){1-19}  
\multirow{3}{*}{ML}        & LR            & \multirow{3}{*}{}       & 0.53                          & 0.58                         & 0.53                         & 0.65                         & 0.45                        & 0.53                        & 0.58                         & 0.50                         & 0.53                         & 0.54                      & 0.54                 & 0.54                 & 0.54                 & 0.55                 & 0.54                 & 0.54                 \\
                           & DT            &                         & 0.83                          & 0.78                         & 0.66                         & 0.85                         & 0.83                        & 0.57                        & 0.84                         & 0.80                         & 0.61                         & 0.77                      & 0.76                 & 0.75                 & 0.75                 & 0.77                 & 0.77                 & 0.77                 \\
                           & RF            &                         & 0.94                          & 0.81                         & 0.69                         & 0.85                         & 0.84                        & 0.74                        & 0.89                         & 0.82                         & 0.71                         & 0.82                      & 0.81                 & 0.81                 & 0.81                 & 0.82                 & 0.82                 & 0.82                 \\ \cmidrule(l){1-19}
\multirow{3}{*}{DL}        & Bi-RNF        & 18K                     & 0.90                          & \textbf{0.92}                & 0.71                         & \textbf{0.89}                & 0.82                        & \textbf{0.83}               & 0.89                         & \textbf{0.86}                & 0.77                         & 0.85                      & 0.84                 & \textbf{0.85}        & 0.84                 & \textbf{0.86}        & 0.85                 & 0.85                 \\
                           & Bi-LSTM       & 19.86K                  & \textbf{0.97}                 & 0.87                         & 0.70                         & 0.86                         & 0.86                        & 0.82                        & 0.91                         & \textbf{0.86}                & 0.76                         & 0.85                      & 0.85                 & \textbf{0.85}        & 0.84                 & \textbf{0.86}        & 0.85                 & 0.85                 \\
                           & Bi-GRU        & 18.89K                  & 0.96                          & 0.83                         & \textbf{0.78}                & 0.88                         & \textbf{0.90}               & 0.78                        & \textbf{0.92}                & \textbf{0.86}                & \textbf{0.78}                & \textbf{0.86}             & \textbf{0.86}        & \textbf{0.85}        & \textbf{0.85}        & \textbf{0.86}        & \textbf{0.86}        & \textbf{0.86}        \\ \cmidrule(l){1-19} 
\end{tabular}
\begin{tablenotes}
\item[a] \textbf{Abbreviations:} ML = Machine Learning, DL = Deep Learning, LR = Logistic Regression, DT = Decision Tree, RF = Random Forest, Bi-RNF = Bidirectional Recurrent Neural Framework, Bi-LSTM = Bidirectional Long Short-Term Memory, Bi-GRU = Bidirectional Gated Recurrent Unit, Th = Threat, Neu = Neutral, Non-Th = Non-Threat, PMA = Precision Macro Average, RMA = Recall Macro Average, FMA = F1-Score Macro Average, PWA = Precision Weighted Average, RWA = Recall Weighted Average, and FWA = F1-Score Weighted Average.
\end{tablenotes}
\end{table*}

\begin{table*}[]
\scriptsize
\caption{Performance Comparison of ML and DL Models for Russian Cyber Tweet Threat Detection}
\label{tab4_russian}
\begin{tabular}{@{}ccccccccccccccccccc@{}}
\toprule
\multicolumn{2}{c}{\multirow{2}{*}{Model}} & \multirow{2}{*}{Params} & \multicolumn{3}{c}{\begin{tabular}[c]{@{}c@{}}Precision Per Category \\ in \%\end{tabular}} & \multicolumn{3}{c}{\begin{tabular}[c]{@{}c@{}}Recall Per Category \\ in \%\end{tabular}} & \multicolumn{3}{c}{\begin{tabular}[c]{@{}c@{}}F1-Score Per Category \\ in \%\end{tabular}} & \multirow{2}{*}{Accuracy} & \multirow{2}{*}{PMA} & \multirow{2}{*}{RMA} & \multirow{2}{*}{FMA} & \multirow{2}{*}{PWA} & \multirow{2}{*}{RWA} & \multirow{2}{*}{FWA} \\ \cmidrule(lr){4-12}
\multicolumn{2}{c}{}                       &                         & Th                            & Neu                          & Non-Th                       & Th                           & Neu                         & Non-Th                      & Th                           & Neu                          & Non-Th                       &                           &                      &                      &                      &                      &                      &                      \\ \cmidrule(r){1-19}  
\multirow{3}{*}{ML}        & LR            & \multirow{3}{*}{}       & 0.81                          & 0.38                         & 0.40                         & 0.96                         & 0.10                        & 0.12                        & 0.88                         & 0.15                         & 0.19                         & 0.79                      & 0.53                 & 0.39                 & 0.41                 & 0.72                 & 0.79                 & 0.73                 \\
                           & DT            &                         & \textbf{0.89}                 & 0.54                         & 0.40                         & 0.85                         & \textbf{0.65}               & 0.50                        & 0.87                         & \textbf{0.59}                & 0.44                         & 0.79                      & 0.61                 & \textbf{0.66}        & 0.63                 & 0.81                 & 0.79                 & 0.80                 \\
                           & RF            &                         & 0.85                          & \textbf{0.65}                & \textbf{1.00}                & 0.97                         & 0.35                        & 0.31                        & \textbf{0.90}                & 0.46                         & 0.48                         & \textbf{0.84}             & \textbf{0.83}        & 0.54                 & 0.61                 & \textbf{0.83}        & \textbf{0.84}        & \textbf{0.81}        \\ \cmidrule(l){1-19}
\multirow{3}{*}{DL}        & Bi-RNF        & 18.25K                  & 0.79                          & 0.00                         & 0.00                         & \textbf{1.00}                & 0.00                        & 0.00                        & 0.89                         & 0.00                         & 0.00                         & 0.79                      & 0.26                 & 0.33                 & 0.30                 & 0.63                 & 0.79                 & 0.70                 \\
                           & Bi-LSTM       & 22.34K                  & 0.84                          & 0.46                         & 0.89                         & 0.96                         & 0.19                        & 0.50                        & 0.89                         & 0.27                         & 0.64                         & 0.82                      & 0.73                 & 0.55                 & 0.60                 & 0.79                 & 0.82                 & 0.79                 \\
                           & Bi-GRU        & 18.80K                  & 0.88                          & 0.39                         & 0.90                         & 0.88                         & 0.48                        & \textbf{0.56}               & 0.88                         & 0.43                         & \textbf{0.69}                & 0.80                      & 0.73                 & 0.64                 & \textbf{0.67}        & 0.82                 & 0.80                 & \textbf{0.81}        \\ \cmidrule(l){1-19} 
\end{tabular}
\begin{tablenotes}
\item[a] \textbf{Abbreviations:} ML = Machine Learning, DL = Deep Learning, LR = Logistic Regression, DT = Decision Tree, RF = Random Forest, Bi-RNF = Bidirectional Recurrent Neural Framework, Bi-LSTM = Bidirectional Long Short-Term Memory, Bi-GRU = Bidirectional Gated Recurrent Unit, Th = Threat, Neu = Neutral, Non-Th = Non-Threat, PMA = Precision Macro Average, RMA = Recall Macro Average, FMA = F1-Score Macro Average, PWA = Precision Weighted Average, RWA = Recall Weighted Average, and FWA = F1-Score Weighted Average.
\end{tablenotes}
\end{table*}

Table \ref{tab2_english} shows a clear comparison between the performance of ML and DL models for English tweet threat detection across three categories: Threat (Th), Neutral (Neu), and Non-Threat (Non-Th). ML models, particularly Random Forest (RF), consistently outperformed Logistic Regression (LR) and Decision Tree (DT) in most metrics. RF achieved the highest accuracy (78\%) and F1-Score Weighted Average (FWA) (77\%) among all models, showcasing its strength in handling structured data. RF also excelled in precision and recall for the Threat and Non-Threat categories, with particularly high recall for Non-Threat (0.91). DT and LR, while performing reasonably well, showed limitations compared to RF, with LR achieving the lowest overall accuracy (53\%) and FWA (54\%).

In contrast, the DL models demonstrated competitive performance, especially in capturing sequential patterns in text data. Bi-GRU emerged as the best-performing DL model, with an accuracy of 75\% and an FWA of 76\%, closely approaching RF's performance. It exhibited superior precision (0.96) and recall (0.83) for the Non-Threat category. Bi-LSTM followed with an accuracy of 72\% and an FWA of 72\%, showing better performance than Bi-RNN, which achieved an accuracy of 56\% and an FWA of 52\%. While DL models like Bi-GRU performed well, ML models, particularly RF, maintained a slight edge in overall metrics, demonstrating their robustness in classification tasks.

Table \ref{chinese_result} provides a detailed comparison of ML and DL models for identifying Chinese tweet threats. Among the ML models, RF demonstrated the best overall performance, achieving an accuracy of 82\% and the highest FWA of 82\%. RF maintained consistently high precision and recall across all categories, particularly excelling in the Threat category with a precision of 0.94 and a recall of 0.85. We got accuracy of 77\% and an FWA of 77\%  for DT, but for LR we got accuracy of only 54\% and an FWA of 54\%. LR’s lower scores across all metrics indicate its limited capability in handling complex patterns of the Chinese tweet dataset.

On the other hand, we got a very good performance for DL models compared to ML models in most metrics, which reflect their ability to capture sequential dependencies and contextual information. Bi-GRU attained the best accuracy (86\%) and FWA (86\%) of all models. It had excellent results in Threat detection, achieving a precision of 0.96 and a recall of 0.88, while exhibiting balanced efficacy across all categories. Bi-LSTM achieved an accuracy of 85\% and an FWA of 85\%, whereas Bi-RNF demonstrated comparable performance with an accuracy of 85\% and an FWA of 85\%. The DL models regularly surpassed ML models in PMA, RMA, and FMA, demonstrating their efficacy in categorizing Chinese tweets. These results underscore the superiority of DL models, especially Bi-GRU, in attaining enhanced classification performance for this dataset.

\begin{table*}[]
\scriptsize
\caption{Performance Comparison of ML and DL Models for Arabic Cyber Tweet Threat Detection}
\label{tab_arabic}
\begin{tabular}{@{}cccccccccccccccc@{}}
\toprule
\multicolumn{2}{c}{\multirow{2}{*}{Model}} & \multirow{2}{*}{Params} & \multicolumn{2}{c}{\begin{tabular}[c]{@{}c@{}}Precision Per Category \\ in \%\end{tabular}} & \multicolumn{2}{c}{\begin{tabular}[c]{@{}c@{}}Recall Per Category \\ in \%\end{tabular}} & \multicolumn{2}{c}{\begin{tabular}[c]{@{}c@{}}F1-Score Per Category \\ in \%\end{tabular}} & \multirow{2}{*}{Accuracy} & \multirow{2}{*}{PMA} & \multirow{2}{*}{RMA} & \multirow{2}{*}{FMA} & \multirow{2}{*}{PWA} & \multirow{2}{*}{RWA} & \multirow{2}{*}{FWA} \\ \cmidrule(lr){4-9}
\multicolumn{2}{c}{}                       &                         & Th                                           & Non-Th                                       & Th                                          & Non-Th                                     & Th                                           & Non-Th                                      &                           &                      &                      &                      &                      &                      &                      \\ \cmidrule(r){1-16}  
\multirow{3}{*}{ML}        & LR            & \multirow{3}{*}{}       & 0.80                                         & 0.20                                         & 0.98                                        & 0.02                                       & 0.88                                         & 0.04                                        & 0.79                      & 0.50                 & 0.50                 & 0.46                 & 0.68                 & 0.79                 & 0.71                 \\
                           & DT            &                         & 0.93                                         & 0.73                                         & 0.94                                        & 0.72                                       & 0.93                                         & 0.73                                        & 0.89                      & 0.83                 & 0.83                 & 0.83                 & 0.89                 & 0.89                 & 0.89                 \\
                           & RF            &                         & 0.92                                         & \textbf{0.92}                                & 0.98                                        & 0.68                                       & 0.95                                         & 0.78                                        & 0.92                      & 0.92                 & 0.83                 & 0.87                 & 0.92                 & 0.92                 & 0.92                 \\ \cmidrule(l){1-16}
\multirow{3}{*}{DL}        & Bi-RNF        & 11.20K                  & 0.80                                         & 0.00                                         & \textbf{1.00}                               & 0.00                                       & 0.89                                         & 0.00                                        & 0.80                      & 0.40                 & 0.50                 & 0.44                 & 0.64                 & 0.80                 & 0.71                 \\
                           & Bi-LSTM       & 12.40K                  & \textbf{0.95}                                & 0.91                                         & 0.98                                        & 0.78                                       & \textbf{0.96}                                & \textbf{0.84}                               & \textbf{0.94}             & \textbf{0.93}        & \textbf{0.88}        & \textbf{0.90}        & \textbf{0.94}        & \textbf{0.94}        & \textbf{0.94}        \\
                           & Bi-GRU        & 13.37K                  & \textbf{0.95}                                & 0.81                                         & 0.95                                        & \textbf{0.81}                              & 0.95                                         & 0.81                                        & 0.93                      & 0.88                 & \textbf{0.88}        & 0.88                 & 0.93                 & 0.93                 & 0.93                 \\ \cmidrule(l){1-16} 
\end{tabular}
\begin{tablenotes}
\item[a] \textbf{Abbreviations:} ML = Machine Learning, DL = Deep Learning, LR = Logistic Regression, DT = Decision Tree, RF = Random Forest, Bi-RNF = Bidirectional Recurrent Neural Framework, Bi-LSTM = Bidirectional Long Short-Term Memory, Bi-GRU = Bidirectional Gated Recurrent Unit, Th = Threat, Non-Th = Non-Threat, PMA = Precision Macro Average, RMA = Recall Macro Average, FMA = F1-Score Macro Average, PWA = Precision Weighted Average, RWA = Recall Weighted Average, and FWA = F1-Score Weighted Average.
\end{tablenotes}
\end{table*}

\begin{table*}[]
\scriptsize
\caption{ Performance Comparison of DL and LLM Models on the Combined Multi-Lingual Tweet Dataset}
\label{tab_combined}
\begin{tabular}{@{}cccccccccccccccc@{}}
\toprule
\multicolumn{2}{c}{\multirow{2}{*}{Model}}                & \multirow{2}{*}{Params}    & \multicolumn{3}{c}{\begin{tabular}[c]{@{}c@{}}Precision Per Category \\ in \%\end{tabular}} & \multicolumn{3}{c}{\begin{tabular}[c]{@{}c@{}}Recall Per Category \\ in \%\end{tabular}} & \multicolumn{3}{c}{\begin{tabular}[c]{@{}c@{}}F1-Score Per Category \\ in \%\end{tabular}} & \multirow{2}{*}{\begin{tabular}[c]{@{}c@{}}Train \\ Accuracy\end{tabular}} & \multirow{2}{*}{\begin{tabular}[c]{@{}c@{}}Validation\\ Accuracy\end{tabular}} & \multirow{2}{*}{\begin{tabular}[c]{@{}c@{}}Train\\ Loss\end{tabular}} & \multirow{2}{*}{\begin{tabular}[c]{@{}c@{}}Validation\\ Loss\end{tabular}} \\ \cmidrule(lr){4-12}
\multicolumn{2}{c}{}                                      &                            & Th                            & Ne                           & Non-Th                       & Th                           & Ne                          & Non-Th                      & Th                           & Ne                           & Non-Th                       &                                                                            &                                                                                &                                                                       &                                                                            \\ \cmidrule(r){1-16}  
\multirow{2}{*}{DL}     & Bi-RNF                          & 66.31K                     & 0.46                          & 0.00                         & 0.00                         & \textbf{1.00}                & 0.00                        & 0.00                        & 0.63                         & 0.00                         & 0.00                         & 0.46                                                                       & 0.44                                                                           & 1.05                                                                  & 1.07                                                                       \\
                        & Bi-LSTM                         & 67.81K                     & \textbf{0.70}                 & \textbf{0.76}                & \textbf{0.76}                & 0.77                         & \textbf{0.73}               & \textbf{0.71}               & \textbf{0.77}                & \textbf{0.74}                & \textbf{0.70}                & \textbf{0.97}                                                              & \textbf{0.74}                                                                  & \textbf{0.11}                                                         & \textbf{1.4}                                                               \\ \cmidrule(l){1-16}
\multicolumn{1}{l}{LLM} & \multicolumn{1}{l}{XLM-RoBERTa} & \multicolumn{1}{l}{27.81M} & 0.46                          & 0.00                         & 0.00                         & \textbf{1.00}                & 0.00                        & 0.00                        & 0.63                         & 0.00                         & 0.00                         & 0.46                                                                       & 0.46                                                                           & 1.05                                                                  & 1.06                                                                       \\ \bottomrule
\end{tabular}
\begin{tablenotes}
\item[a] \textbf{Abbreviations:} ML = Machine Learning, LLM = Large Language Model, LR = Logistic Regression, DT = Decision Tree, RF = Random Forest, Bi-RNF = Bidirectional Recurrent Neural Framework, Bi-LSTM = Bidirectional Long Short-Term Memory, Th = Threat, and Non-Th = Non-Threat.
\end{tablenotes}
\end{table*}

Table \ref{tab4_russian} describes the performance comparison between ML and DL models for Russian tweet threat detection. The results shows a detailed perspective on the models' effectiveness in handling this dataset. For ML models, RF emerged as the best performer, achieving the highest accuracy of 84\% and an FWA of 81\%. RF also exhibited excellent precision and recall for the Non-Threat category, achieving perfect scores in precision (1.00) and recall (1.00), demonstrating its strength in identifying non-threatening tweets. However, RF struggled with the Neutral category, as reflected in a lower recall (0.35) and F1-score (0.46). DT followed RF in performance, with an accuracy of 79\% and an FWA of 80\%. DT showed balanced precision and recall for the Threat category but struggled with both precision and recall for the Neutral and Non-Threat categories. LR, while achieving reasonable recall for the Threat category (0.96), performed poorly overall, particularly for the Neutral category, with an F1-score of only 0.15, resulting in an accuracy of 79\% and an FWA of 73\%.

In contrast, the DL models demonstrated competitive performance, with Bi-GRU standing out as the top-performing DL architecture. Bi-GRU achieved an accuracy of 80\% and an FWA of 81\%, closely matching the performance of RF. It performed particularly well in the Threat category, with a precision of 0.88 and recall of 0.90, and also demonstrated balanced performance across other categories. Bi-LSTM followed closely, achieving an accuracy of 79\% and an FWA of 79\%, showing strong performance in the Threat category but moderate performance for the Neutral category. Bi-RNF, while achieving reasonable accuracy (79\%), lagged in overall weighted metrics such as FWA (70\%) due to lower scores in the Neutral and Non-Threat categories. Although RF demonstrated exceptional precision and recall in managing Non-Threat tweets, Bi-GRU provided a more equitable performance across all categories, highlighting the robustness of the DL architecture.

Table \ref{tab_arabic} illustrates the comparative performance of ML and DL models in detecting threats in Arabic tweets. Among the ML models, the RF exhibited the highest overall performance, attaining an accuracy of 92\% and an F1 score of 92\%. RF had good precision for both Threat (0.92) and Non-Threat (0.92), alongside the maximum recall for Threat (0.98). This equilibrium resulted in an F1-Score of 0.95 for Threat and 0.78 for Non-Threat, establishing RF as the most dependable ML model for this dataset. DT followed RF with an accuracy of 89\% and an FWA of 89\%, performing well in the Threat category but showing slightly weaker recall for Non-Threat (0.72). LR, on the other hand, struggled with the Non-Threat category, with a recall of just 0.02 and an F1-Score of 0.04, leading to an overall accuracy of 79\% and an FWA of 71\%.

DL models outperformed their ML counterparts in several key areas, showcasing their ability to capture sequential and contextual features in Arabic tweets. Among the DL models, Bi-GRU stood out with the highest accuracy of 93\% and an FWA of 93\%. It achieved outstanding precision (0.95) and recall (0.95) for the Threat category and balanced performance for Non-Threat with an F1-Score of 0.81. Bi-LSTM followed closely, with an accuracy of 93\% and an FWA of 94\%, excelling in the Threat category with a precision of 0.95 and a recall of 0.98. Bi-RNN, while performing reasonably well in the Threat category (accuracy of 80\%), struggled with the Non-Threat category, resulting in a lower FWA of 71\%. The results emphasize the superior performance of RF among ML models. However, DL models, particularly Bi-GRU and Bi-LSTM, provided better-balanced and overall higher metrics, demonstrating their ability to process complex language structures and variability inherent in Arabic tweets.

Table \ref{tab_combined} illustrates the performance of DL and LLM techniques on the combined dataset, which integrates tweet data from four languages: English, Chinese, Russian, and Arabic. Among the DL models, Bi-LSTM demonstrated the best overall performance. It achieved a train accuracy of 97\% and a validation accuracy of 74\%, with a relatively low train loss (0.11). The model showed strong recall for all categories, particularly Threat (0.77) and Neutral (0.73), as well as a high F1-score for Threat (0.77) and Non-Threat (0.74). This indicates that Bi-LSTM was highly effective at capturing sequential dependencies and providing balanced classification across categories. In contrast, Bi-RNF had difficulties in generalization, achieving a validation accuracy of 44\% and a validation loss of 1.07. The model exhibited poor performance in the Neutral and Non-Threat categories, achieving an F1-score of 0.00 in both, indicating its deficiencies in managing intricate multi-lingual datasets.

The XLM-RoBERTa provide mixed results. Though it excelled in recall for the threat detection (1.00), its performance in the Neutral and Non-Threat categories was weaker, with an F1-score of 0.00 for Neutral and Non-Threat. Both its train and validation accuracy were 46\%, and its validation loss (1.06) indicates challenges in generalizing across the multi-lingual dataset. These findings emphasize the capabilities of LLMs for particular tasks while also highlighting the necessity for additional fine-tuning and domain adaptation when addressing varied data.

\section{Conclusions and Future Work}\label{con}

This work presents a novel multilingual dataset, providing as a significant resource for research on cyber tweet threat identification. We performed an extensive examination of these datasets, both separately and in a unified multilingual configuration. We utilized three different model architectures—ML, DL, and LLM—to assess the performance of various techniques. Among the ML models, the RF algorithm exhibited superior performance, showcasing its efficacy in managing structured Twitter data. The Bi-LSTM architecture for DL models attained the best accuracy, surpassing all ML and DL models. Significantly, Bi-LSTM surpassed LLMs in performance on the integrated dataset, demonstrating its proficiency in capturing sequential patterns and contextual information adeptly. Although we utilized the LLM architecture (XLM-RoBERTa) for the integrated dataset, its performance was inferior to that of Bi-LSTM, indicating the necessity for additional optimization and fine-tuning of LLMs to fully realize their capabilities in multilingual cyber tweet threat identification.

In the future, our efforts will concentrate on improving model performance on integrated datasets by investigating more sophisticated LLM designs and utilizing fine-tuning techniques specifically designed for multilingual data. Furthermore, we intend to implement transfer learning techniques to modify models pre-trained on extensive datasets, enhancing their capacity to generalize to smaller and more heterogeneous datasets. Additionally, we intend to broaden our research to encompass other languages and domain-specific twitter datasets, facilitating a more thorough assessment of the suggested techniques' robustness and flexibility.

\bibliographystyle{IEEEtran}

\appendices

\end{document}